# Online Factorization and Partition of Complex Networks From Random Walks

Lin F. Yang, Vladimir Braverman, Tuo Zhao, Mengdi Wang *

**Abstract**

Finding the reduced-dimensional structure is critical to understanding complex networks. Existing approaches such as spectral clustering are applicable only when the full network is explicitly observed. In this paper, we focus on the online factorization and partition of implicit large-scale networks based on observations from an associated random walk. We formulate this into a nonconvex stochastic factorization problem and propose an efficient and scalable stochastic generalized Hebbian algorithm. The algorithm is able to process dependent state-transition data dynamically generated by the underlying network and learn a low-dimensional representation for each vertex. By applying a diffusion approximation analysis, we show that the continuous-time limiting process of the stochastic algorithm converges globally to the "principal components" of the Markov chain and achieves a nearly optimal sample complexity. Once given the learned low-dimensional representations, we further apply clustering techniques to recover the network partition. We show that when the associated Markov process is lumpable, one can recover the partition exactly with high probability. We apply the proposed approach to model the traffic flow of Manhattan as city-wide random walks. By using our algorithm to analyze the taxi trip data, we discover a latent partition of the Manhattan city that closely matches the traffic dynamics.

*Lin Yang and Mengdi Wang are affiliated with Department of Operations Research and Financial Engineering, Princeton University. Vladimir Braverman is affiliated with Department of Computer Science at Johns Hopkins University; Tuo Zhao is affiliated with School of Industrial and Systems Engineering at Georgia Institute of Technology; Email: lin.yang@princeton.edu, mengdiw@princeton.edu; Mengdi Wang is the corresponding author. This material is based upon work supported by the NSF Grants IIS-1447639, EAGER CCF- 1650041, and CAREER CCF-1652257.

# 1 Introduction

Network data arise in many applications and research areas, including but not limited to social science, economics, transportation, finance, power grid, artificial intelligence, etc. Examples include protein-protein interaction networks [12], phone communication networks [21], collaboration networks [4], and the gravitational interaction network of dark matter particles in cosmology [23, 15, 20]. Due to the highly complex nature of these networks, many efforts have been devoted to investigating their reduced-order representations from high-dimensional data (e.g. [7, 24, 22, 5]).

In this paper, we focus on learning from the dynamic "state-transition" data, which are snapshots of a random walk associated with the implicit network. For example, records of taxi trips can be used to reveal the traffic dynamics of a metropolitan. Each trip can be viewed as a fragmented sample path realized from a city-wide Markov chain that characterizes the traffic dynamics [14, 3]. None of the existing works has considered how to recover the latent network partition of an urban area from the taxi trip data. For another example, reinforcement learning applications such as autonomous driving and game AI are modeled as Markov decision processes [26], which unfortunately suffer from the curse of dimensionality of the state space. Given trajectories of game snapshots or a game simulator, it is of vital interest to identify the low-dimensional representation of the "state" of game. For the general problem of finding reduced-order representations, popular approaches such as principal component analysis and spectral clustering do not utilize the Markov nature of state-transition data. Existing computational methods often require explicit knowledge and pre-computation of large matrices, which cannot scale to large-scale problems and is not even possible for online learning applications. Efficient methods are in demand.

Motivated by the need to analyze state-transition data, we propose an efficient and scalable approach for online factorization and partition of implicit complex networks. We start by employing a stochastic gradient-type algorithm, namely the generalized Hebbian algorithm (GHA), and tailor it towards processing Markov transition data. Then we show that the GHA learns low-dimensional representations of the network in an online fashion, and by further applying clustering techniques, we can recover the underlying partition structure with high probability. Our analysis is based on a diffusion approximation approach, which is widely used in stochastic analysis of complicated discrete processes such as queueing networks (see [10] for more related literature on diffusion approximation). By properly rescaling of time, we approximate the discrete-time dynamics generated by the GHA algorithm using its continuous-time limiting process, which is the solution to an ordinary differential equation (ODE). Though the stochastic optimization problem is highly nonconvex, we show that the limiting stochastic process of the GHA converges geometrically to the global optima, even if the initial solution is chosen uni-formly at random. We further show that the process after sufficiently large time is well approximated by an Ornstein-Uhlenbeck process, whose stochastic fluctuation can be precisely characterized. Despite of the spherical geometry and many unstable equilibria of the optimization problem, we establish global convergence with a near-optimal sample complexity guarantee in an asymptotic manner.

Our work is partly motivated by [29], which establishes the connection between networks and a class of *lumpable* Markov chains. It proposes an optimization framework to identify the partition structure when the transition matrix is known *a priori*. Our method is also related to the class of online eigenvalue decomposition methods for representation learning [1, 13, 30, 2, 11]. However, none of the existing methods and analysis are applicable to Markov transition data and online network partition.

**Notation:** We denote $[n] = \{1, 2, \ldots, n\}$. Given two matrices $\boldsymbol{U} \in \mathbb{R}^{m\times r_1}, \boldsymbol{V} \in \mathbb{R}^{m\times r_2}$ with orthonormal columns, where $1 \le r_1 \le r_2 \le m$, we denote the principle angle between two matrices by $\Theta(\boldsymbol{U}, \boldsymbol{V}) = \mathrm{diag}\Big[\cos^{-1}(\sigma_1(\boldsymbol{U}^\top \boldsymbol{V})),\ \cos^{-1}(\sigma_2(\boldsymbol{U}^\top \boldsymbol{V})),\ \ldots,\ \cos^{-1}\big(\sigma_{r_1}(\boldsymbol{U}^\top \boldsymbol{V})\big)\Big]$, where $\sigma_i(\boldsymbol{A})$ is the $i$-th largest singular value of matrix $\boldsymbol{A}$. We also use $\cos(\cdot)$ and $\sin(\cdot)$ to act on matrices and denote entry-wise functions. For a matrix $\boldsymbol{V}$, we denote by $\boldsymbol{V}_{*j}$ its $j$-th column vector and by $\boldsymbol{V}_{i*}$ its $i$-th row vector. We denote by $\boldsymbol{V}_{*1:r}$ the sub-matrix of the first $r$ columns. We denote by $\|\cdot\|_{\mathrm{F}}$ the Frobenius norm of a matrix, and denote by $\|\cdot\|_2$ the Euclidean norm of a vector or the spectral norm of a matrix. We denote by $\boldsymbol{e}_i \in \mathbb{R}^s$ the $i$-th standard unit vector for any $s \ge i$: $(\boldsymbol{e}_i)_i = 1$ and $(\boldsymbol{e}_i)_j = 0$ for $j \ne i$. We also denote by $0_{m\times n} \in \mathbb{R}^{m\times n}$ the matrix with all 0 entries.

# 2 Preliminaries

Let us review the basics of networks and the associated Markov chains.

**Networks and Associated Markov Chains:** Let $G = (S, E)$ be a *connected* network with $m$ vertices (a weighted directed graph), where $S = \{s_1, s_2, \ldots, s_m\}$ denotes the vertex set, $E = \big\{w_{i,j} \ge 0 : i, j \in [m]\big\}$ denotes the edge set, and $w_{i,j}$ denotes the weight of the edge $(s_i, s_j)$.

Consider the random walk that is naturally associated with the network $G$: We denote by $\boldsymbol{P} = \big(p_{i,j}\big) \in \mathbb{R}^{m\times m}$ its probability transition matrix, where each state of the Markov chain corresponds to a vertex in $G$. Since $G$ is a connected network, all states of the Markov chain are recurrent. The Markov chain generated by the network $G$ satisfies $\mathbb{P}\Big[s^{(t)} = s_j \Big| s^{(t-1)} = s_i\Big] = p_{i,j}$. Suppose that $G$ is *undirected* (i.e., $w_{ij} = w_{ji}$), then $\forall i, j : p_{i,j} = \frac{w_{i,j}}{w_i}$ and $w_i = \sum_{j\in[m]} w_{i,j}$. The stationary distribution of the Markov chain is $\mu_i = \frac{w_i}{\sum_{j\in[m]} w_j}$. The corresponding Markov chain is *reversible* and satisfies the following *detailed balance*

*condition*

$$\forall i \neq j,\ \mu_i p_{i,j} = \mu_j p_{j,i} \quad \text{and} \quad \sum_{i \in [m]} \mu_i p_{i,j} = \mu_j, \tag{1}$$

i.e., $\boldsymbol{DP} = \boldsymbol{PD}$, where $\boldsymbol{D} = \mathrm{diag}(\mu_1, \mu_2, \ldots, \mu_m)$. Note that our subsequent analysis does not require the undirectedness assumption of the underlying network. In this paper, we focus on connected and undirected networks where $\mu_i > 0$ for all $i \in [m]$. For a non-connected network, our method still applies with the caveat that it recovers the structure of a connected component determined by the initial state.

**Our Problem of Interest** Given a sample trajectory $\{s^{(0)}, s^{(1)}, \ldots, s^{(t)}, \ldots\}$ of state transitions of the unknown Markov chain, our objective is to develop an online learning method to extract reduced-order information about the Markov chain and recover the latent network partition.

We are interested in complex networks that can be approximated using reduced-order representations. To be general, we consider networks with associated Markov chains *nearly low-rank*, which is defined as follows:

**Definition 1** (Nearly Low-Rank Markov Chains). *A Markov chain with transition matrix* $\boldsymbol{P}$ *is nearly low-rank if there exist matrices* $\boldsymbol{F}_1, \boldsymbol{F}_2 \in \mathbb{R}^{m\times m}$, *where* $\mathrm{rank}(\boldsymbol{F}_1) = r$ *and* $\|\boldsymbol{F}_2\|_2 < \sigma_r(\boldsymbol{F}_1)$ *such that*

$$\boldsymbol{DP} = \boldsymbol{F}_1 + \boldsymbol{F}_2 \quad \textit{and} \quad \boldsymbol{F}_1^\top \boldsymbol{F}_2 = 0_{m\times m}, \tag{2}$$

*and* $\boldsymbol{F}_1 = \boldsymbol{U}\Sigma\boldsymbol{V}^\top$, *where* $\Sigma = \mathrm{diag}(\sigma_1, \sigma_2, \ldots, \sigma_r)$ *is a diagonal matrix with* $1 \geq \sigma_1 \geq \sigma_2 \geq \ldots \geq \sigma_r > 0$, *and* $\boldsymbol{U}, \boldsymbol{V} \in \mathbb{R}^{m\times r}$ *are matrices with orthonormal columns.*

Consider the following *representation matrix*

$$\boldsymbol{M} := \boldsymbol{D}^{-1}\boldsymbol{V} \in \mathbb{R}^{m\times r}, \tag{3}$$

each row of which can be viewed as an $r$-dimensional representation of a vertex of $G$. The matrix $\boldsymbol{M}$ gives a set of approximate "principal components" of the Markov chain, which has a similar spirit as spectral clustering [7]. Note that Markov chains that are nearly low-rank are not necessarily reversible. When a Markov chain is both nearly low-rank and reversible, the conditions in Definition 1 shall hold with $\boldsymbol{U} = \boldsymbol{V}$.

In particular, we also consider an important special case of nearly low-rank Markov chains - "lumpable" Markov chains, which is introduced by [16] and formally in [29] as follows.

**Definition 2** (Special Case: Lumpable Markov chains [29]). *A reversible Markov chain on states* $S$ *with transition matrix* $\boldsymbol{P}$ *is* lumpable *with respect to the partition* $S = S_1 \cup S_2 \ldots \cup S_r$ *if the top* $r$ *eigenvectors of* $\boldsymbol{DP}$ *are piecewise constant with respect to the* $S_1, \ldots, S_r$.

We can view $S_1,\ldots,S_r$ as"meta states" of the Markov chain. When the lumpability condition holds, the transitions between these sets satisfy the strong Markov property, i.e., for any $s_k, s_h \in S_i$, $\forall\ j$,

$$\sum_{s_\ell \in S_j} p_{k,\ell} = \sum_{s_\ell \in S_j} p_{h,\ell}.$$

Intuitively speaking, the meta states suffice to characterize the macro dynamics of a complex Markov chain. When the Markov chain is lumpable, it is nearly-low rank as in Definition 1 with $\boldsymbol{U} = \boldsymbol{V}$. In this case, the matrix $\boldsymbol{U}$ becomes a block matrix. For any $i, j \in [r]$, the vector $\boldsymbol{U}_{*i}$ restricted on coordinates $S_j$ has constant values across all entries. The work [29] showed when the Markov chain is lumpable with respect to a partition $S = S_1 \cup S_2 \ldots \cup S_r$, one can recover the exact partition by clustering its $r$-dimensional representations (rows of $\boldsymbol{M} = \boldsymbol{D}^{-1}\boldsymbol{V}$). An example of a network and its lumpable Markov chain is given in Section 5.1.

# 3 Method

Recall that we are interested in learning from Markov transition data. In particular, consider the scenario where we only observe state-to-state transitions of a Markov process over $S$: $s^{(1)}, s^{(2)}, s^{(3)}, \ldots, s^{(n-1)}, s^{(n)}, \ldots$, without knowing the transition matrix $\boldsymbol{P}$ in advance. For notational convenience, we simplify the notation of the states to $S = \{1, 2, \ldots m\}$.

## 3.1 A Nonconvex Optimization Model for Markov Chain Factorization

To handle the dependency of the Markov process, we need to downsample the data. Specifically, we divide the trajectory of $n$ state transitions into $b$ blocks with block size $\tau$ for some $\tau \geq 2$:

$$\underbrace{s^{(1)}, s^{(2)}, \ldots, s^{(\tau)}}_{\text{the 1-st block}}, \underbrace{s^{(\tau+1)}, s^{(\tau+2)}, \ldots, s^{(2\tau)}}_{\text{the 2-nd block}}, \ldots, \underbrace{s^{(b-1)\tau+1}, s^{(b-1)\tau+2}, \ldots s^{(b\tau)}}_{\text{the b-th block}}.$$

For the $k$-th block, we select the last two samples and construct $\boldsymbol{Z}^{(k)} \in \mathbb{R}^{m\times m}$ to be the matrix with one entry equaling 1 and all other entries equaling 0, i.e.,

$$\boldsymbol{Z}^{(k)}_{s^{(k\tau-1)}, s^{(k\tau)}} = 1 \quad \text{and} \quad \boldsymbol{Z}^{(k)}_{s,s'} = 0 \text{ for all } (s, s') \neq \left(s^{(k\tau-1)}, s^{(k\tau)}\right). \tag{4}$$

Here we choose a large enough $\tau$ such that $\forall k \geq 1$, $\mathbb{E}\left[\boldsymbol{Z}^{(k)} \middle| s^{(0)}\right] \approx \boldsymbol{D}\boldsymbol{P} = \boldsymbol{F}_1 + \boldsymbol{F}_2$,where $\boldsymbol{F}_1 = \boldsymbol{U}^\top \boldsymbol{\Sigma} \boldsymbol{V}$ and $\boldsymbol{F}_2$ are given in Definition 1. Intuitively, the choice of $\tau$ shall be related to how fast the Markov chain mixes. We will specify the choice of $\tau$ in Section 4.

Let us formulate the Stochastic Transition Matrix Decomposition Problem as

$$(\boldsymbol{U}^*, \boldsymbol{V}^*) = \underset{\widetilde{\boldsymbol{U}}, \widetilde{\boldsymbol{V}} \in \mathbb{R}^{m\times r}}{\text{argmax}} \ \text{tr}\left[\widetilde{\boldsymbol{U}}^\top \mathbb{E}\boldsymbol{Z}\widetilde{\boldsymbol{V}}\right] \quad \text{subject to} \quad \widetilde{\boldsymbol{U}}^\top\widetilde{\boldsymbol{U}} = \widetilde{\boldsymbol{V}}^\top\widetilde{\boldsymbol{V}} = I_r. \tag{5}$$

where the expectation

$$\mathbb{E}\boldsymbol{Z} := \lim_{n\to\infty} n^{-1}\sum_{k=1}^{n} \boldsymbol{Z}^{(k)} = \boldsymbol{D}\boldsymbol{P}$$

is taken over the invariant distribution of the Markov chain. Note that $\boldsymbol{U}^*$ and $\boldsymbol{V}^*$ are global optima to (5), and they satisfy $\boldsymbol{U}^* = \boldsymbol{U}\boldsymbol{O}$ and $\boldsymbol{V}^* = \boldsymbol{V}\boldsymbol{O}$ for some orthonormal matrix $\boldsymbol{O} \in \mathbb{R}^{r\times r}$. By using a self-adjoint dilation, we recast (5) into a symmetric decomposition problem as follows

$$\boldsymbol{W}^* = \underset{\boldsymbol{W}\in\mathbb{R}^{2m\times r}}{\text{argmax}}\, \text{tr}\left[\boldsymbol{W}^\top \mathbb{E}\boldsymbol{A}\boldsymbol{W}\right] \quad \text{subject to} \quad \boldsymbol{W}^\top\boldsymbol{W} = I_r, \tag{6}$$

where $\mathbb{E}\boldsymbol{A} = \begin{bmatrix} 0_{m\times m} & \mathbb{E}\boldsymbol{Z} \\ \mathbb{E}\boldsymbol{Z}^\top & 0_{m\times m} \end{bmatrix} \in \mathbb{R}^{2m\times 2m}$ and $\boldsymbol{W} = \frac{1}{\sqrt{2}}[\boldsymbol{U}^\top, \boldsymbol{V}^\top]^\top \in \mathbb{R}^{2m\times r}$.

## 3.2 Algorithm for Online Factorization of Markov Chains

To solve (6), we adopt the Generalized Hebbian Algorithm (GHA) which was originally developed for training neural nets and principal component analysis [25]. GHA, also referred as Sanger's rule, is essentially a stochastic primal-dual algorithm. Specifically, let $\mathcal{L}(\boldsymbol{W}, \boldsymbol{L})$ be the Lagrangian function of Eq. (6) given by

$$\mathcal{L}(\boldsymbol{W}, \boldsymbol{L}) = \text{tr}\left[\boldsymbol{W}^\top \mathbb{E}\boldsymbol{A}\boldsymbol{W}\right] - \text{tr}\left[\boldsymbol{L}\left(\boldsymbol{W}^\top\boldsymbol{W} - \boldsymbol{I}_r\right)\right],$$

where $\boldsymbol{L} \in \mathbb{R}^{r\times r}$ is the Lagrangian multiplier matrix. By checking the Karush-Kuhn-Tucker (KKT) conditions of the problem $\max_{\boldsymbol{W}} \min_{\boldsymbol{L}} \mathcal{L}(\boldsymbol{W}, \boldsymbol{L})$, we obtain

$$\mathbb{E}\boldsymbol{A}\boldsymbol{W}^* + \boldsymbol{W}^*\boldsymbol{L}^* = 0 \quad \text{and} \quad \boldsymbol{W}^{*\top}\boldsymbol{W}^* - \boldsymbol{I}_r = 0, \tag{7}$$

where $\boldsymbol{L}^*$ is the optimal Lagrangian multiplier. The above KKT conditions further imply

$$\boldsymbol{L}^* = -\boldsymbol{W}^{*\top}\mathbb{E}\boldsymbol{A}\boldsymbol{W}^*. \tag{8}$$

GHA is essentially an stochastic approximation method for the solving the equations (7) and (8). Specifically, we use the $k$-th block of transition data to compute the sample matrix

$$\boldsymbol{A}^{(k)} = \begin{bmatrix} 0_{m\times m} & \boldsymbol{Z}^{(k)} \\ \boldsymbol{Z}^{(k)\top} & 0_{m\times m} \end{bmatrix} \in \mathbb{R}^{2m\times 2m}. \tag{9}$$

Then the $k$-th iteration of GHA takes the form

$$\text{Dual Update} \quad : \quad \boldsymbol{L}^{(k)} = \underbrace{\boldsymbol{W}^{(k)\top}\boldsymbol{A}^{(k)}\boldsymbol{W}^{(k)}}_{\text{Markov sample of } \boldsymbol{W}^{(k)\top}\mathbb{E}\boldsymbol{A}\boldsymbol{W}^{(k)}} \tag{10}$$

$$\text{Primal Update} \quad : \quad \boldsymbol{W}^{(k+1)} = \boldsymbol{W}^{(k)} + \eta \underbrace{(\boldsymbol{A}^{(k)}\boldsymbol{W}^{(k)} - \boldsymbol{W}^{(k)}\boldsymbol{L}^{(k)})}_{\text{Markov sample of } \nabla_{\boldsymbol{W}}\mathcal{L}(\boldsymbol{W}^{(k)},\Lambda^{(k)})} \tag{11}$$

where $\eta > 0$ is the learning rate. Combing (10) with (11), we get a dual-free update of GHA as follows,

$$\boldsymbol{W}^{(k+1)} = \boldsymbol{W}^{(k)} + \eta(\boldsymbol{A}^{(k)}\boldsymbol{W}^{(k)} - \boldsymbol{W}^{(k)}\boldsymbol{W}^{(k)\top}\boldsymbol{A}^{(k)}\boldsymbol{W}^{(k)}).$$

Note that the columns of $\boldsymbol{W}^{(k)}$ are not necessarily orthogonal. But when $\boldsymbol{W}^{(0)}$ has orthonormal columns, then $\boldsymbol{W}^{(k)}$ tends to have orthonormal columns as $\eta \to 0$. The formal procedure is presented in Algorithm 1.

**Algorithm 1** SGA for Online Factorization of Markov Chains

**Output**: A stream of Markov transition data $s^{(1)}, s^{(2)}, s^{(3)}, \ldots, s^{(n-1)}, s^{(n)}, \ldots$
**Initialize:**
 Sample matrix $G \in \mathbb{R}^{2m\times r}$ with i.i.d. entries from $\mathcal{N}(0,1)$;
 $\boldsymbol{W}^{(0)} \leftarrow QR(G)$, $k \leftarrow 0$;
**Repeat:**
For every $\tau$ state transitions, obtain $\boldsymbol{A}^{(k)}$ using Eqs. (4),(9);
 $\boldsymbol{W}^{(k+1)} \leftarrow \boldsymbol{W}^{(k)} + \eta\left[\boldsymbol{A}^{(k)}\boldsymbol{W}^{(k)} - \boldsymbol{W}^{(k)}\boldsymbol{W}^{(k)\top}\boldsymbol{A}^{(k)}\boldsymbol{W}^{(k)}\right]$;
 $k \leftarrow k+1$;
**Until** stopping condition is satisfied
**Output** $[\widehat{\boldsymbol{U}}; \widehat{\boldsymbol{V}}] \leftarrow \sqrt{2}\boldsymbol{W}^{(k)}$

Algorithm 1 is a globally convergent method which does not require any warm-up initialization or prior knowledge. The initial solution $\boldsymbol{W}^{(0)}$ is drawn uniformly from the set of all orthonormal matrices by applying a QR decomposition to a matrix with i.i.d. Gaussian entries. Algorithm 1 makes update online and uses $O(mr)$ space, while a batch method needs $O(m^2)$ space to store the explicit transition matrix.

## 3.3 Recovering The Network Partition from Random Walks

Recall that in Definition 1 the $m \times r$ matrix $\boldsymbol{M} = \boldsymbol{D}^{-1}\boldsymbol{V}$ gives a reduced-order representation for each vertex of the network. As long as we can estimate $\boldsymbol{D}, \boldsymbol{V}$, we would be able to partition the network by applying a clustering algorithm such as the $k$-means. Let us describe the overall procedure:

**(1)** Run Algorithm 1 on the Markov transition data and obtain $[\widehat{\boldsymbol{U}};\widehat{\boldsymbol{V}}]$.
**(2)** Let $\widehat{\boldsymbol{\mu}}$ be the empirical estimate of the stationary distribution, i.e., $\widehat{\mu}_i = \sum_{k=1}^n \mathbb{I}(s^{(k)} = i)/n$. Let $\widehat{\boldsymbol{D}} = \mathrm{diag}(\widehat{\mu}_1, \widehat{\mu}_2, \ldots, \widehat{\mu}_m)$. Now each row of $\widehat{\boldsymbol{M}} = \widehat{\boldsymbol{D}}^{-1}\widehat{\boldsymbol{V}}$ gives an approximate $r$-dimensional representation for the corresponding state/vertex.
**(3)** Find a set of centers $C = \{c_1, c_2, \ldots, c_r\} \subset \mathbb{R}^r$ by solving the following problem:

$$\widehat{C} = \operatorname*{argmin}_C \sum_{i=1}^m \min_{c \in C} d^2(\widehat{\boldsymbol{M}}_{i*}, c), \tag{12}$$

where $d(\widehat{\boldsymbol{M}}_{s_i*}, c_j) = \|\widehat{\boldsymbol{M}}_{s_i*} - c_j\|_2$ is the Euclidean distance.
**(4)** Output the partition by assigning each state to its closest center.

# 4 Theory

In order to show the convergence of Algorithm 1, we uses the idea of diffusion approximation (e.g. [10]). (1) We show that the dynamics of our algorithm can be approximated by an ordinary differential equation (ODE); (2) To analyze the convergence rate, we show that after proper rescaling of time, the algorithm's dynamics can be characterized by the solution of a Stochastic Difference Equation (SDE). The SDE allows us to analyze the error fluctuation when the iterates are within a small neighborhood of the global optimum.

## 4.1 Reducing Dependency by Down Sampling

Recall our goal is to estimate Markov chain factorization from random walks. In the online learning setting, the data comes in a stream and are highly independent. To handle the dependency, our algorithm replies on down sampling the data points. Next we introduce some important measures for a Markov chain. For notational convenience, we denote $\mu(\Omega) = \sum_{i\in\Omega} \mu_i$ for any subset of states $\Omega \subset S$. We introduce the *merging conductance* [18] of a Markov chain by

$$\Phi = \min_{\Omega \subset [m]} \frac{\sum_{j\in\Omega, \ell\in\Omega^c} \sum_{i\in[m]} \frac{\mu_j p_{j,i} \mu_\ell p_{\ell,i}}{\mu_i}}{\sum_{j\in\Omega} \mu_j} \quad \text{subject to} \quad \mu(\Omega) \leq 1/2,$$

where $\Omega^c$ is the complement of $\Omega$. The parameter $\Phi$ is a generalization of the Cheeger's constant, which characterizes the bottleneck of a network. For recurrent Markov chains that are rapidly mixing, $\Phi$ can be treated as a constant. Besides, we let

$$\mu_{\max} = \max_{i\in[m]} \mu_i, \qquad \mu_{\min} = \min_{i\in[m]} \mu_i.$$

We then choose block length $\tau$ in Algorithm 1 as follows:

$$\tau \geq \left[\frac{2}{\Phi^2} \log\left(\sqrt{\frac{\mu_{\max}}{\mu_{\min}}} \frac{1}{\eta}\right)\right].$$

As we will shown in section A.3, by choosing such a down-sampling block length, our data samples are sufficiently close to i.i.d. samples drawn from the stationary distribution of the underlying Markov chain. This allows us to approximate algorithm by another auxiliary procedure of fully independent samples. In the next two sub-sections, we show the limiting procedure of the algorithm based on this down-sampling rate.

## 4.2 ODE Characterization of Algorithm 1

Let $\boldsymbol{R} \in \mathbb{R}^{2m\times 2m}$ be the matrix of eigenvectors of $\mathbb{E}\boldsymbol{A}$ in (6). We consider a transformation by $\boldsymbol{R}$:

$$\overline{\boldsymbol{W}}^{(k)} = \boldsymbol{R}\boldsymbol{W}^{(k)} \quad \text{and} \quad \overline{\boldsymbol{A}}^{(k)} = \boldsymbol{R}^\top \boldsymbol{A}^{(k)}\boldsymbol{R}.$$

Let $\boldsymbol{\Lambda} = \mathrm{diag}(\sigma_1, \sigma_2, \ldots, \sigma_{2m}) = \mathbb{E}\overline{\boldsymbol{A}}$ with that $\sigma_1 \geq \sigma_2 \geq \ldots \sigma_{2m}$. To demonstrate an ODE characterization for the trajectory of the algorithm, we introduce a continuous time $t$. Recall where $\eta$ is the learning rate. We denote $\overline{\boldsymbol{W}}(t) = \overline{\boldsymbol{W}}^{(\lfloor t/\eta \rfloor)}$. For notation simplicity, we may drop $(t)$ if it is clear from the context. For $r+1 \leq i \leq 2m$, we define the cosine subspace angle as

$$\gamma_i^{(\eta)}(t) = \left\| \boldsymbol{e}_i^\top \overline{\boldsymbol{R}}^\top \boldsymbol{W} \right\|_2 = \left\| \boldsymbol{e}_i^\top \overline{\boldsymbol{W}} \right\|_2,$$

where $\boldsymbol{e}_i \in \mathbb{R}^{2m}$ is the $i$-th standard unit vector. We use $(\eta)$ as a superscript to emphasize the dependence on $\eta$. To show a global convergence of $\gamma_i^{(\eta)}(t)$, we characterize its upper bound in the following lemma.

**Lemma 1** (Principle Angle Upper Bound). *Let $\boldsymbol{E} = (\boldsymbol{e}_1, \boldsymbol{e}_2, \ldots, \boldsymbol{e}_r) \in \mathbb{R}^{2m\times r}$. Suppose that $\boldsymbol{W}$ has orthonormal columns and $\boldsymbol{E}^\top \boldsymbol{W}$ is full rank. For any $\boldsymbol{X} \in \mathbb{R}^{2m\times s}$ with $s \geq 1$, we have $\|\boldsymbol{X}^\top \boldsymbol{W}\|_{\mathrm{F}} \leq \left\|\boldsymbol{X}^\top \boldsymbol{W} \cdot (\boldsymbol{E}^\top \boldsymbol{W})^{-1}\right\|_{\mathrm{F}}$.*

Accordingly, we define

$$\widetilde{\gamma}_i^{(\eta)} = \left\| \boldsymbol{e}_i^\top \overline{\boldsymbol{W}} \cdot \left(\boldsymbol{E}^\top \overline{\boldsymbol{W}}\right)^{-1} \right\|_2.$$

Since $\boldsymbol{W}^{(0)}$ has orthonormal columns, for any fixed $t > 0$, the columns of $\overline{\boldsymbol{W}}(t)$ are orthonormal almost surely as $\eta \to 0$. Thus $\widetilde{\gamma}_i^{(\eta)}$ becomes a uniform upper bound of $\gamma_i^{(\eta)}$ almost surely as $\eta \to 0$. The next theorem establishes the continuous time limit for $\widetilde{\gamma}_i^{(\eta)}$.

**Theorem 1** (ODE Convergence). *Given $\overline{\boldsymbol{W}}^{(0)}$ with orthonormal columns and that $\boldsymbol{E}^\top \overline{\boldsymbol{W}}^{(0)}$ is invertible, for all $r < i \leq 2m$, $\widetilde{\gamma}_i^{(\eta)}(t)$ converges weakly to the solution of the following ODE,*

$$\frac{d\widetilde{\gamma}_i^2(t)}{dt} = b_i \widetilde{\gamma}_i^2(t)$$

*as $\eta \to 0$, where $b_i$ is some constant satisfying $b_i \leq 2(\sigma_i - \sigma_r)$.*

Theorem 1 suggests the global convergence of the algorithm. Specifically, the solution to the above ODE is

$$\widetilde{\gamma}_i(t) = \widetilde{\gamma}_i(0)e^{b_i t/2} \leq \widetilde{\gamma}_i(0)e^{(\sigma_i-\sigma_r)t}, \qquad \forall\ r < i \leq 2m,$$

which implies $\gamma_i^{(\eta)}(t) \to 0$ for any $r < i \leq 2m$ as $\eta \to 0$ and $t \to \infty$. Since $\left\|\sin\Theta\left(\boldsymbol{E}, \overline{\boldsymbol{W}}(t)\right)\right\|_{\mathrm{F}}^2 = \sum_{i>r} \gamma_i^{(\eta)2}(t)$, we obtain

$$\begin{aligned}\left\|\sin\Theta\left(\widehat{\boldsymbol{U}}(t), \overline{\boldsymbol{U}}\right)\right\|_{\mathrm{F}}^2 + \left\|\sin\Theta\left(\widehat{\boldsymbol{V}}(t), \overline{\boldsymbol{V}}\right)\right\|_{\mathrm{F}}^2 &\leq 2\left\|\sin\Theta\left(\boldsymbol{E}, \overline{\boldsymbol{W}}(t)\right)\right\|_{\mathrm{F}}^2 \\ &\leq 2\sum_{i>r} \widetilde{\gamma}_i(0)e^{(\sigma_{r+1}-\sigma_r)t} \to 0. \end{aligned} \tag{13}$$

## 4.3 SDE Characterization of Algorithm 1

Our ODE approximation of the algorithm shows that after sufficiently many iterations with sufficiently small $\eta$, the algorithm solution can be arbitrarily close to the true subspace, $\mathrm{span}(\overline{\boldsymbol{R}}_{*1}, \overline{\boldsymbol{R}}_{*2}, \ldots, \overline{\boldsymbol{R}}_{*r})$. To obtain the "rate of convergence", however, we need to study the variance of the trajectory at time $t$. Note that such a variance is of order $\mathcal{O}(\eta)$, and vanishing under the limit of $\eta \to 0$. To characterize the variance, we need to rescale the updates by a factor of $\eta^{-1/2}$, i.e., after rescaling, the variance is of order $\mathcal{O}(1)$. Specifically, the rescaled update is defined as

$$\zeta_i^{(\eta)}(t) = \eta^{-1/2} \cdot \overline{\boldsymbol{W}}^{(\lfloor t/\eta \rfloor)\top} \cdot \boldsymbol{e}_i \in \mathbb{R}^r.$$

Note that given $\boldsymbol{W}^{(k)}$ such that $\mathrm{span}(\boldsymbol{W}^{(k)}) = \mathrm{span}(\overline{\boldsymbol{R}}_{*1}, \overline{\boldsymbol{R}}_{*2}, \ldots, \overline{\boldsymbol{R}}_{*r})$, we have

$$\mathbb{E}\left(\boldsymbol{W}^{(k+1)} | \boldsymbol{W}^{(k)}\right) = \boldsymbol{W}^{(k)}.$$

Namely, any matrix in $\mathrm{span}(\overline{\boldsymbol{R}}_{*1}, \overline{\boldsymbol{R}}_{*2}, \ldots, \overline{\boldsymbol{R}}_{*r})$ is a fixed point for Equation (19), in expectation. We consider a regime, where the algorithm has already run for sufficient many iterations such that

$$\left\|\sin\Theta\left(\boldsymbol{R}_{*1:r}, \boldsymbol{W}^{(N_1)}\right)\right\|_{\mathrm{F}}^2 = \left\|\sin\Theta\left(\boldsymbol{E}, \overline{\boldsymbol{W}}^{(N_1)}\right)\right\|_{\mathrm{F}}^2 \leq \eta^{1/c},$$

for some constant $c > 1$. By restarting the counter, we denote $\overline{\boldsymbol{W}}^{(0)} := \overline{\boldsymbol{R}}^\top \boldsymbol{W}^{(N_1)}$. Now we define $\overline{\boldsymbol{W}}^{(0)\top} \boldsymbol{\Lambda} \overline{\boldsymbol{W}}^{(0)} = \boldsymbol{\Gamma}^\top \cdot \widetilde{\boldsymbol{\Lambda}}_r \cdot \boldsymbol{\Gamma}$, where $\boldsymbol{\Gamma} \in \mathbb{R}^{r\times r}$ is an orthonormal matrix and $\widetilde{\boldsymbol{\Lambda}}_r = \mathrm{diag}\left(\sigma_1', \sigma_2', \ldots, \sigma_r'\right)$, with $\sigma_1' \geq \sigma_2' \geq \ldots \geq \sigma_r' \geq 0$.

Denote $\zeta_{i,j}^{(\eta)}(t) = \eta^{-1/2}\left(\boldsymbol{e}_j'^\top \boldsymbol{\Gamma} \cdot \overline{\boldsymbol{W}}^{(\lfloor t/\eta \rfloor)\top} \cdot \boldsymbol{e}_i\right)$, for $i = r+1, r+2, \ldots, 2m$ and $j = 1, 2, \ldots, r$, where $\boldsymbol{e}_j' \in \mathbb{R}^r$ denotes the $j$-th standard unit vector in $\mathbb{R}^r$. We establish the following theorem.

**Theorem 2** (SDE Convergence). *Given* $\left\|\sin\Theta\left(\boldsymbol{E},\overline{\boldsymbol{W}}^{(\lfloor t/\eta\rfloor)}\right)\right\|_{\mathrm{F}}^2 \leq \mathcal{O}(\eta^{1/c})$ *for all* $t \geq 0$, *then for any* $i > r$ *and* $j \in [r]$, *the trajectory of* $\zeta_{i,j}^{(\eta)}(t)$ *weakly converges to the solution of the following SDE, as* $\eta \to 0$,

$$d\zeta_{i,j} = K_{i,j}\zeta_{i,j}dt + G_{i,j}d\mathcal{B}_{i,j} \tag{14}$$

*where* $\mathcal{B}_{i,j}$ *is the standard Brownian motion (not necessarily i.i.d. across* $i,j$*) and constants* $K_{i,j} \leq (\sigma_i - \sigma_r)$, $\sum_{i>r}^{2m} G_{i,j}^2 \leq B \quad$ *for any* $\quad j \in [r]$, *with some absolute constant* $B$.

Notice that (14) is a Fokker-Plank equation, which admits the following solution,

$$\zeta_{i,j}(t) = \zeta_{i,j}(0)\exp\left[K_{i,j}t\right] + G_{i,j}\int_0^t \exp\left[K_{i,j}(s-t)\right]d\mathcal{B}_{i,j}(s). \tag{15}$$

Therefore, we show that each $\zeta_{i,j}^{(\eta)}(t)$ weakly converges to an Ornstein-Uhlenbeck (OU) process, which is widely studied in existing literature [17]. Since the drifting term is driven by $K_{i,j} < 0$, the OU process eventually becomes a pure random walk, i.e., the first term of R.H.S. in (15) goes to 0. Recall that $\zeta_{i,j}^{(\eta)}(t)$ characterizes the sin angle of the subspaces, i.e.,

$$\left\|\sin\Theta\left(\boldsymbol{E},\overline{\boldsymbol{W}}^{(\lfloor t/\eta\rfloor)}\right)\right\|_{\mathrm{F}}^2 = \eta\sum_{i>r}^{2m}\sum_{j=1}^{r}\zeta_{i,j}^{(\eta)2}(t). \tag{16}$$

Thus the fluctuation of $\zeta_{i,j}^{(\eta)}(t)$ is essentially the error fluctuation of the algorithm after sufficiently many iterations. By (15), we obtain

$$\begin{aligned}\mathbb{E}\left\|\sin\Theta\left(\boldsymbol{E},\overline{\boldsymbol{W}}^{(\lfloor t/\eta\rfloor)}\right)\right\|_{\mathrm{F}}^2 &\asymp \eta\sum_{i>r}^{2m}\sum_{j=1}^{r}G_{i,j}^2\int_0^t \exp\left[2K_{i,j}t\right]dt \\ &= \mathcal{O}\left(\frac{\eta r}{\left|K_{i,j}\right|}\right) = \mathcal{O}\left(\frac{\eta r}{\sigma_r(\boldsymbol{F}_1) - \|\boldsymbol{F}_2\|_2}\right).\end{aligned}$$

Given the error parameter $\epsilon > 0$, we need $\eta$ to satisfy $\mathcal{O}\left(\frac{\eta r}{\sigma_r(\boldsymbol{F}_1)-\|\boldsymbol{F}_2\|_2}\right) \asymp \epsilon$. Combining with a Markov inequality and Equation (13), we obtain the following lemma,

**Lemma 2** (Error Analysis of the Limiting Process). *Given a sufficiently small* $\epsilon > 0$, *let*

$$N = \mathcal{O}\left(\frac{rB}{\epsilon\,(\sigma_r(\boldsymbol{F}_1) - \|\boldsymbol{F}_2\|_2)^2}\log\frac{\sum_{i>r}\widetilde{\gamma}_i^2(0)\cdot B}{\epsilon\,(\sigma_r(\boldsymbol{F}_1) - \|\boldsymbol{F}_2\|_2)}\right) \quad \text{and} \quad t = N\eta.$$

*Let* $[\widehat{\boldsymbol{U}}(t), \widehat{\boldsymbol{V}}(t)] \leftarrow \boldsymbol{W}(t)$. *We then have*

$$\lim_{\epsilon\to 0}\mathbb{P}\left[\left\|\sin\Theta\left(\widehat{\boldsymbol{U}}(t),\boldsymbol{U}\right)\right\|_{\mathrm{F}}^2 + \left\|\sin\Theta\left(\widehat{\boldsymbol{V}}(t),\boldsymbol{V}\right)\right\|_{\mathrm{F}}^2 > \epsilon\right] \leq \frac{1}{10}.$$

**Remark 1.** *With standard characterizations of the random matrices (e.g. [27]), we obtain the value of* $\sum_{i>r} \widetilde{\gamma}_i^2(0) = \mathrm{poly}(m)$ *with probability close to 1 when m is large. If the ODE and SDE faithfully approximates the algorithm at sufficiently small* $\eta$*, i.e., the approximation error of ODE/SDE to the algorithm updates is smaller than the desired precision* $\epsilon$*, then the number of down-sampling steps of the algorithm is*

$$N = \mathcal{O}\left(\frac{rB}{\epsilon\,(\sigma_r(\boldsymbol{F}_1) - \|\boldsymbol{F}_2\|_2)^2} \log \frac{\sum_{i>r} \widetilde{\gamma}_i^2(0) \cdot B}{\epsilon\,(\sigma_r(\boldsymbol{F}_1) - \|\boldsymbol{F}_2\|_2)}\right).$$

*Our numerical experiments verify this sample complexity empirically in Section 5.*

## 4.4 Recovery of Network Partition By Clustering

We have been established bounds for obtaining sate embeddings in the last two sections. Next we show that one can recover the partition structure of the underlying network, provided that the Markov chain is lumpable and the sample size is sufficiently large.

**Theorem 3** (Recovery of Partition Structure for Lumpable Markov Chains)**.** *Suppose that the estimated eigen-matrices* $\widehat{\boldsymbol{U}}$*,* $\widehat{\boldsymbol{V}}$*, and empirical distribution* $\widehat{\boldsymbol{\mu}}$ *satisfy*

$$\|\sin\Theta(\widehat{\boldsymbol{U}},\boldsymbol{U})\|_F^2 + \|\sin\Theta(\widehat{\boldsymbol{V}},\boldsymbol{V})\|_F^2 \le \epsilon \text{ and } \max_{i\in[m]} \left|\widehat{\mu}_i - \mu_i\right| \le \sqrt{\epsilon}\mu_i. \tag{17}$$

*for some* $\epsilon \in (0,1)$*. Let* $\widehat{\boldsymbol{M}} := \mathrm{diag}(\widehat{\boldsymbol{\mu}})^{-1}\widehat{\boldsymbol{V}}$ *and* $\boldsymbol{M}$ *as defined in Definition 1. Then for any* $s_i, s_j \in S$*,*

$$\left|\left\|\widehat{\boldsymbol{M}}_{s_i*} - \widehat{\boldsymbol{M}}_{s_j*}\right\|_2^2 - \left\|\boldsymbol{M}_{s_i*} - \boldsymbol{M}_{s_j*}\right\|_2^2\right| \le \frac{C\epsilon}{\mu_{\min}^2}.$$

*Moreover, suppose that the Markov chain is lumpable with respect to the partition* $S_1, \ldots, S_r$*. Then the procedure of Section 3.3 exactly recovers the network partition as long as*

$$\forall l, s_i \in S_l, s_j \in S_l^c : \left\|\boldsymbol{M}_{s_i*} - \boldsymbol{M}_{s_j*}\right\|_2^2 > \frac{2C\epsilon}{\mu_{\min}^2}.$$

Theorem 3 implies that our proposed partition approach can exactly recover the partition of a lumpable Markov chain, as long as the random walk trajectory is long enough to tell the blocks apart. It is possible to extend our analysis to approximately lumpable Markov chains, which is left for future research. The proof is given in Section A.4.

# 5 Experiments

We experiment with the proposed method on two data sets.

## 5.1 Simulated Data

Consider a 12-vertex lumpable network that has 3 meta-states (see Figure 1(a)). We let its probability transition matrix be $\boldsymbol{P}_{\text{ex}}$ (which is specified in (18)). We generate the Markov transition data by simulating the random walk according to $\boldsymbol{P}_{\text{ex}}$. We test our algorithms on the simulated transition data using $10^4$ samples for 100 independent trials. In every single trial, the meta states are correctly recovered. Figure 1(b) shows that the convergence rate of subspace angle are consistent with Remark 1. For comparison, we test a convex relaxation algorithm (MSG [2]) for online solution of Problem (5) (there is no theoretic guarantee for applying MSG to depend data). Figure 1(c) shows that our algorithm significantly outperforms MSG in wall-clock time complexity using both fixed and diminishing stepsizes.

$$\boldsymbol{P}_{\text{ex}} = \\ \begin{bmatrix}
0 & \frac{84}{625} & \frac{463}{10000} & \frac{101}{1250} & \frac{463}{10000} & \frac{84}{625} & \frac{463}{10000} & \frac{101}{1250} & \frac{101}{1250} & \frac{84}{625} & \frac{101}{1250} & \frac{84}{625} \\
\frac{323}{2000} & 0 & \frac{323}{2000} & \frac{17}{250} & \frac{323}{2000} & \frac{273}{10000} & \frac{323}{2000} & \frac{17}{250} & \frac{17}{250} & \frac{273}{10000} & \frac{17}{250} & \frac{273}{10000} \\
\frac{463}{10000} & \frac{84}{625} & 0 & \frac{101}{1250} & \frac{463}{10000} & \frac{84}{625} & \frac{463}{10000} & \frac{101}{1250} & \frac{101}{1250} & \frac{84}{625} & \frac{101}{1250} & \frac{84}{625} \\
\frac{147}{1000} & \frac{103}{1000} & \frac{147}{1000} & 0 & \frac{147}{1000} & \frac{103}{1000} & \frac{147}{1000} & 0 & 0 & \frac{103}{1000} & 0 & \frac{103}{1000} \\
\frac{463}{10000} & \frac{84}{625} & \frac{463}{10000} & \frac{101}{1250} & 0 & \frac{84}{625} & \frac{463}{10000} & \frac{101}{1250} & \frac{101}{1250} & \frac{84}{625} & \frac{101}{1250} & \frac{84}{625} \\
\frac{323}{2000} & \frac{273}{10000} & \frac{323}{2000} & \frac{17}{250} & \frac{323}{2000} & 0 & \frac{323}{2000} & \frac{17}{250} & \frac{17}{250} & \frac{273}{10000} & \frac{17}{250} & \frac{273}{10000} \\
\frac{463}{10000} & \frac{84}{625} & \frac{463}{10000} & \frac{101}{1250} & \frac{463}{10000} & \frac{84}{625} & 0 & \frac{101}{1250} & \frac{101}{1250} & \frac{84}{625} & \frac{101}{1250} & \frac{84}{625} \\
\frac{147}{1000} & \frac{103}{1000} & \frac{147}{1000} & 0 & \frac{147}{1000} & \frac{103}{1000} & \frac{147}{1000} & 0 & 0 & \frac{103}{1000} & 0 & \frac{103}{1000} \\
\frac{147}{1000} & \frac{103}{1000} & \frac{147}{1000} & 0 & \frac{147}{1000} & \frac{103}{1000} & \frac{147}{1000} & 0 & 0 & \frac{103}{1000} & 0 & \frac{103}{1000} \\
\frac{323}{2000} & \frac{273}{10000} & \frac{323}{2000} & \frac{17}{250} & \frac{323}{2000} & \frac{273}{10000} & \frac{323}{2000} & \frac{17}{250} & \frac{17}{250} & 0 & \frac{17}{250} & \frac{273}{10000} \\
\frac{147}{1000} & \frac{103}{1000} & \frac{147}{1000} & 0 & \frac{147}{1000} & \frac{103}{1000} & \frac{147}{1000} & 0 & 0 & \frac{103}{1000} & 0 & \frac{103}{1000} \\
\frac{323}{2000} & \frac{273}{10000} & \frac{323}{2000} & \frac{17}{250} & \frac{323}{2000} & \frac{273}{10000} & \frac{323}{2000} & \frac{17}{250} & \frac{17}{250} & \frac{273}{10000} & \frac{17}{250} & 0
\end{bmatrix}. \tag{18}$$

The transition matrix on the three meta states is,

$$\boldsymbol{P}_r = \begin{bmatrix} 0 & 0.5880 & 0.4120 \\ 0.3233 & 0.1389 & 0.5378 \\ 0.2720 & 0.6461 & 0.0819 \end{bmatrix}.$$

The stationary distribution of $\boldsymbol{P}_{\text{ex}}$ is

$$\boldsymbol{\mu}_{\text{ex}} = \{0.105, 0.0874, 0.105, 0.0577, 0.105, 0.0874, \\ 0.105, 0.0577, 0.0577, 0.0874, 0.0577, 0.0874\},$$

and the merging conductance is roughly 0.06. Using $10^4$ sample state transitions, we obtain a fairly close estimate of stationary distribution given by

$$\widehat{\boldsymbol{\mu}}_{\text{ex}} = \{0.106, 0.088, 0.107, 0.057, 0.102, 0.087, \\ 0.105, 0.059, 0.057, 0.087, 0.058, 0.087\}.$$

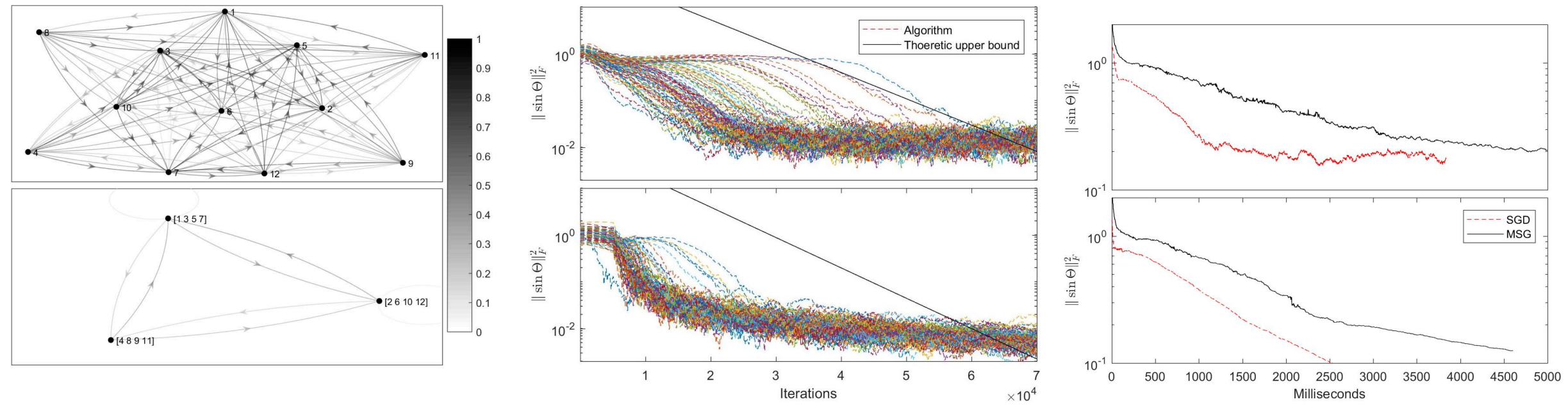

(a) Lumpble Markov Chain (b) Dynamics of SGD (c) SGD v.s. MSG

Figure 1: (a) An illustration of lumpable Markov chain: the full network (Top) and 3 meta-states in the simplified network (Bottom); (b) The convergence in subspace angle of 100 simulations: fixed stepsize (Top) and diminishing step size (Bottom); (c) Time complexity, nonconvex SGD v.s. convex MSG: fixed stepsize (Top) and diminishing step size (Bottom).

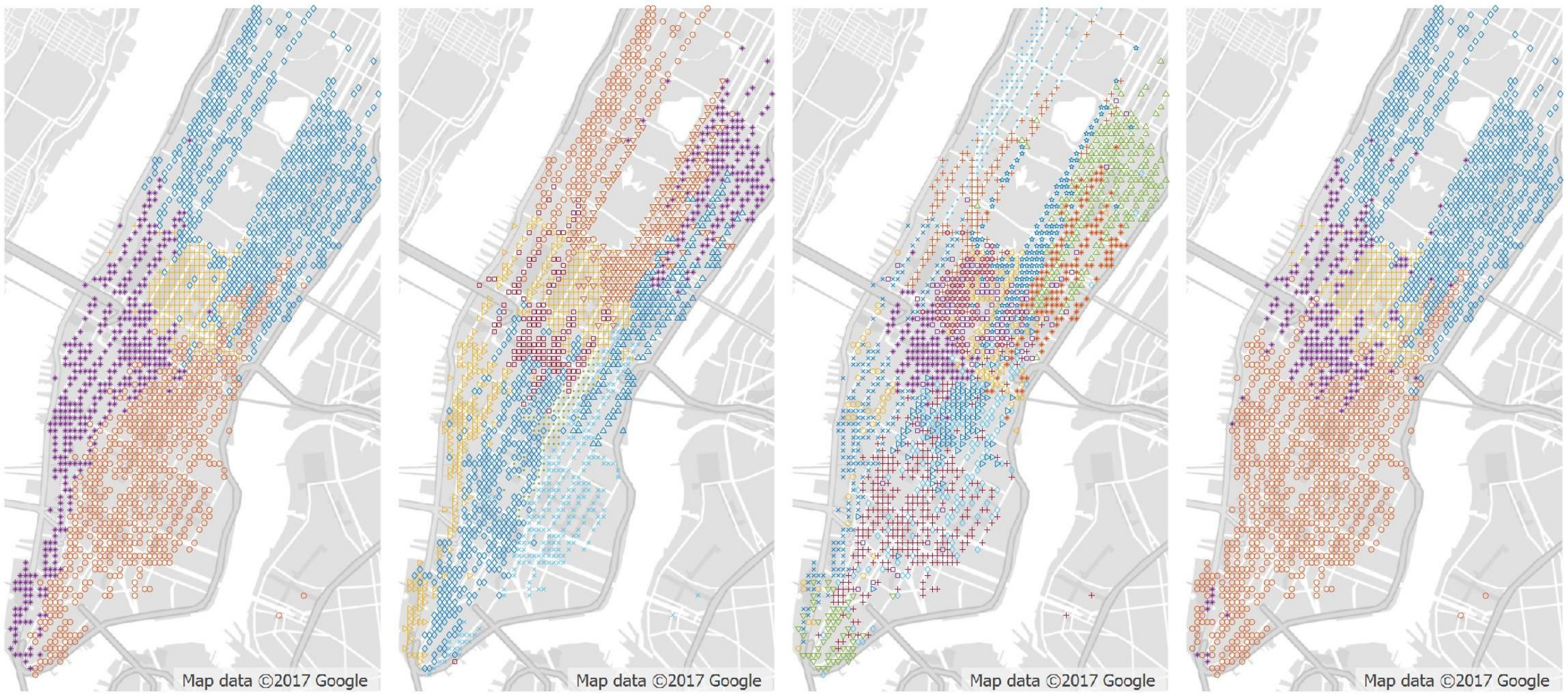

(a) Online: 4 states; (b) Online: 10 states; (c) Online: 15 states; (d) Batch: 4 states.

Figure 2: The meta-states partition of Manhattan traffic network based on taxi trip records. Each color or symbol represents a meta-state.

## 5.2 Manhattan Taxi Data

We experiment using a real dataset that contains $1.1 \times 10^7$ trip records of NYC Yellow cabs from January 2016 [28]. Each entry records the coordinates of the pick-up and drop-off locations, distance and length of trip, and taxifares. We discretize the map into a fine grid (with cell size roughly 10m) and model each taxi trip as a single state transition of a Markov chain. For example, a taxi picks up a customer at cell $s_1$ and drops off the customer at cell $s_2$, then picks up a customer at $s_3$... We can view $s_1, s_2, s_3 \ldots$, as the path of states visited by an implicit city-wide random walk. In order to guarantee recurrence of the random walk, we removed cells that are rarely visited. We end up with 2017 locations and a total of $10^7$ effective trips. We apply Algorithm 1 and the partition procedure of Section 3.3 to the taxi trip data and illustrate the results in Figure 2(a)–2(c). Our method reveals a very informative partition of the Manhattan city according to traffic dynamics. We compare our online algorithm with a batch partition procedure and observe that they generate highly similar results (Figure 2(d)) when $r = 4$. A practically impressive observation is: our algorithm uses less than 1 Mbytes memory for $r = 4, 10, 15$. In contrast, the batch partition uses about 200 Mbytes memory even for $r = 4$.

# 6 Conclusions and Discussions

We have developed an online learning method for analyzing dynamic transition data generated by a random walk on a network. Out method finds the low-dimensional representation of an implicit network and reveals its latent partition structure. Our method has superior space and computation complexity. We show that it achieves near-optimal global convergence and sample complexity by using an ODE-SDE argument.

Our algorithm and analysis can be adapted to work for the more general online singular value decomposition poblem, which has been considered by [11, 13, 1, 30, 2, 6]. The most critical distinction between our result and existing ones is that ours applies to random walk data and network partition. We summarize other technical improvements of our results:

**(1)** All the existing analyses require i.i.d samples, while ours applies to dependent Markov samples; **(2)** [11, 13, 1, 6] analyzed Oja's algorithm for PCA problem, which conducts QR factorization in each iteration. Our algorithm does not require such decomposition - each iteration uses only vector-to-vector inner products; **(3)** [30] analyzed a similar algorithm as ours but their results require a sufficiently near-optimal initial solution, which is not available in our problems. **(4)** [2] investigated a method based on convex relaxation of SVD but they achieves a sub-optimal sample complexity - $\widetilde{O}(1/\epsilon^2)$. Moreover, their algorithm needs to compute an expensive Fantope projection with a computational complexity $O(m^3)$ per iteration.

In summary, Markov transition data carries rich information about the underlying structure of complex networks and stochastic systems. We hope this work will motivate more research in this area and faster algorithms for applications in networks and reinforcement learning.

## References

[1] Allen-Zhu, Z. and Li, Y. (2016). First efficient convergence for streaming k-pca: a global, gap-free, and near-optimal rate. *arXiv preprint arXiv:1607.07837*.

[2] Arora, R., Mianjy, P., and Marinov, T. (2016). Stochastic optimization for multiview representation learning using partial least squares. In *Proceedings of The 33rd International Conference on Machine Learning*. 1786–1794.

[3] Benson, A. R., Gleich, D. F., and Lim, L.-H. (2017). The spacey random walk: A stochastic process for higher-order data. *SIAM Review* **59**, 2, 321–345.

[4] Blondel, V. D., Guillaume, J.-L., Lambiotte, R., and Lefebvre, E. (2008). Fast unfolding of communities in large networks. *Journal of statistical mechanics: theory and experiment* **2008**, 10, P10008.

[5] Chen, J. and Yuan, B. (2006). Detecting functional modules in the yeast protein–protein interaction network. *Bioinformatics* **22**, 18, 2283–2290.

[6] Chen, Z., Yang, L. F., Li, C. J., and Zhao, T. (2017). Online partial least square optimization: Dropping convexity for better efficiency and scalability. 777–786.

[7] Chung, F. R. (1997). *Spectral graph theory*. Vol. **92**. American Mathematical Soc.

[8] Cohen, M. B., Lee, Y. T., Miller, G., Pachocki, J., and Sidford, A. (2016). Geometric median in nearly linear time. In *Proceedings of the 48th Annual ACM SIGACT Symposium on Theory of Computing*. ACM, 9–21.

[9] Ethier, S. N. and Kurtz, T. G. (2009). *Markov processes: characterization and convergence*. Vol. **282**. John Wiley & Sons.

[10] Harold, J., Kushner, G., and Yin, G. (1997). Stochastic approximation and recursive algorithm and applications. *Application of Mathematics 35*.

[11] Jain, P., Jin, C., Kakade, S. M., Netrapalli, P., and Sidford, A. (2016). Streaming pca: Matching matrix bernstein and near-optimal finite sample guarantees for ojas algorithm. In *29th Annual Conference on Learning Theory*. 1147–1164.

[12] Junker, B. H. and Schreiber, F. (2011). *Analysis of biological networks*. Vol. **2**. John Wiley & Sons.

[13] Li, C. J., Wang, M., Liu, H., and Zhang, T. (2016). Near-optimal stochastic approximation for online principal component estimation. *arXiv preprint arXiv:1603.05305*.

[14] Liu, Y., Kang, C., Gao, S., Xiao, Y., and Tian, Y. (2012). Understanding intra-urban trip patterns from taxi trajectory data. *Journal of geographical systems* **14**, 4, 463–483.

[15] Mantegna, R. N. (1999). Hierarchical structure in financial markets. *The European Physical Journal B-Condensed Matter and Complex Systems* **11**, 1, 193–197.

[16] Meila, M. and Shi, J. (2001). A random walks view of spectral segmentation.

[17] Meucci, A. (2009). Review of statistical arbitrage, cointegration, and multivariate ornstein-uhlenbeck.

[18] Mihail, M. (1989). Conductance and convergence of markov chains-a combinatorial treatment of expanders. In *Foundations of computer science, 1989., 30th annual symposium on*. IEEE, 526–531.

[19] Minsker, S. and others. (2015). Geometric median and robust estimation in banach spaces. *Bernoulli* **21**, 4, 2308–2335.

[20] Nerur, S., Sikora, R., Mangalaraj, G., and Balijepally, V. (2005). Assessing the relative influence of journals in a citation network. *Communications of the ACM* **48**, 11, 71–74.

[21] Newman, M. E. (2001). The structure of scientific collaboration networks. *Proceedings of the National Academy of Sciences* **98**, 2, 404–409.

[22] Page, L., Brin, S., Motwani, R., and Winograd, T. (1999). The pagerank citation ranking: Bringing order to the web. Tech. rep., Stanford InfoLab.

[23] Peebles, P. J. E. (1980). *The large-scale structure of the universe*. Princeton university press.

[24] Pizzuti, C. (2008). Ga-net: A genetic algorithm for community detection in social networks. *Parallel problem solving from nature–PPSN X*, 1081–1090.

[25] Sanger, T. D. (1989). Optimal unsupervised learning in a single-layer linear feedforward neural network. *Neural networks* **2**, 6, 459–473.

[26] Sutton, R. S. and Barto, A. G. (1998). *Reinforcement learning: An introduction*. Vol. **1**. MIT press Cambridge.

[27] Tao, T. (2012). *Topics in random matrix theory*. Vol. **132**. American Mathematical Society Providence, RI.

[28] TLC, N. (2017). Nyc taxi and limousine commission (tlc) trip record data. `http://www.nyc.gov/html/tlc/html/about/trip_record_data.shtml`.

[29] Weinan, E., Li, T., and Vanden-Eijnden, E. (2008). Optimal partition and effective dynamics of complex networks. *Proceedings of the National Academy of Sciences* **105**, 23, 7907–7912.

[30] Xie, B., Liang, Y., and Song, L. (2015). Scale up nonlinear component analysis with doubly stochastic gradients. In *Advances in Neural Information Processing Systems*. 2341–2349.

# A Proof Details

## A.1 Proof of Theorem 1

We first assume that the sample $\boldsymbol{A}^{(k)}$s are i.i.d. to each other. In particular, we assume $s^{(i)}$ is generated from the stationary distribution $\mu$ and $s^{(i+1)}$ is generated from distribution $\boldsymbol{P}_{s^{(i)}*}$. Later on, we will remove this requirement. We denote $\mathcal{D}$ as the distribution for the i.i.d samples described above. It can be verified that $\mathcal{D}$ satisfies the following two properties.

**Assumption 1** (Subgaussian moments)**.** *Suppose $\boldsymbol{A} \sim \mathcal{D}$. Then for each $r \in \mathbb{N}_+$,*

$$\mathbb{E}\boldsymbol{A} = \Sigma \quad \textit{and} \quad \mathbb{E}\|(\boldsymbol{A})^r\|_2 \leq C_1^r,$$

*for some constants $C_1$ dependents only on $m$.*

For our distribution $\mathcal{D}$, we observe that

$$\Sigma = \begin{bmatrix} 0_{m\times m} & \boldsymbol{D}\boldsymbol{P} \\ \boldsymbol{P}^\top \boldsymbol{D}^\top & 0_{m\times m} \end{bmatrix} \in \mathbb{R}^{2m\times 2m}.$$

**Assumption 2** (Rotational Invariant Variance)**.** *Suppose $\boldsymbol{A} \sim \mathcal{D}$. Then for any set of orthonormal vectors $\boldsymbol{v}_1, \boldsymbol{v}_2, \ldots, \boldsymbol{v}_{2m}$,*

$$\sum_{j=2}^{2m} \mathbb{E}\left(\boldsymbol{v}_1^\top \boldsymbol{A}\boldsymbol{v}_j\right)^2 \leq B,$$

*for some constant $B$.*

We also introduce the following notations,

$$\boldsymbol{E}_r = (\boldsymbol{e}_1, \boldsymbol{e}_2, \ldots, \boldsymbol{e}_r) \in \mathbb{R}^{2m\times r} \quad \text{and} \quad \overline{\boldsymbol{E}} = (\boldsymbol{e}_{r+1}, \boldsymbol{e}_{r+2}, \ldots, \boldsymbol{e}_{2m}) \in \mathbb{R}^{2m\times(2m-r)}$$

where $\boldsymbol{e}_i$ is the $i$-th standard unit vector in $\mathbb{R}^m$. Let

$$\Sigma = \overline{\boldsymbol{R}}\Lambda\overline{\boldsymbol{R}}^\top$$

be the eigenvalue decomposition of $\Sigma$, i.e., $\overline{\boldsymbol{R}} = (\boldsymbol{v}_1, \boldsymbol{v}_2, \dots, \boldsymbol{v}_{2m})$ are the eigenvectors of $\Sigma$ and $\boldsymbol{\Lambda} = \mathrm{diag}(\sigma_1, \sigma_2, \dots, \sigma_m, -\sigma_m, -\sigma_{m-1}, \dots, -\sigma_1)$. Initialize with a random matrix $\boldsymbol{W}^{(0)} \in \mathbb{R}^{2m\times r}$ with orthonormal columns. Let $\eta > 0$ be the choice of the learning rate. The update for step $k$ is

$$\boldsymbol{W}^{(k+1)} \leftarrow \boldsymbol{W}^{(k)} + \eta \left( \boldsymbol{A}^{(k)} \boldsymbol{W}^{(k)} - \boldsymbol{W}^{(k)} \cdot \boldsymbol{W}^{(k)\top} \boldsymbol{A}^{(k)} \boldsymbol{W}^{(k)} \right).$$

Let $\overline{\boldsymbol{W}} = \boldsymbol{R}^\top \boldsymbol{W}$ and $\overline{\boldsymbol{A}} = \boldsymbol{R}^\top \boldsymbol{A} \boldsymbol{R}$, we obtain,

$$\overline{\boldsymbol{W}}^{(k+1)} \leftarrow \overline{\boldsymbol{W}}^{(k)} + \eta \left( \overline{\boldsymbol{A}}^{(k)} \overline{\boldsymbol{W}}^{(k)} - \overline{\boldsymbol{W}}^{(k)} \cdot \overline{\boldsymbol{W}}^{(k)\top} \overline{\boldsymbol{A}}^{(k)} \overline{\boldsymbol{W}}^{(k)} \right). \tag{19}$$

*Proof of Lemma 1.* Since $\boldsymbol{E}$ and $\boldsymbol{W}$ have orthonormal rows, we have that $\|\boldsymbol{E}^\top \boldsymbol{W}\|_2 \le 1$. Then we immediately have

$$\begin{aligned}
\left\| \boldsymbol{X}^\top \boldsymbol{W} \right\|_{\mathrm{F}} &\le \left\| \boldsymbol{X}^\top \boldsymbol{W} \cdot \left( \boldsymbol{E}^\top \boldsymbol{W} \right)^{-1} \left( \boldsymbol{E}^\top \boldsymbol{W} \right) \right\|_{\mathrm{F}} \\
&\le \left\| \boldsymbol{X}^\top \boldsymbol{W} \cdot \left( \boldsymbol{E}^\top \boldsymbol{W} \right)^{-1} \right\|_{\mathrm{F}} \left\| \boldsymbol{E}^\top \boldsymbol{W} \right\|_2,
\end{aligned}$$

as desired. □

*Proof of Lemma 1 (On Independent Sample).* First we note that if $\boldsymbol{E}^\top \overline{\boldsymbol{W}}^{(k)}$ is invertible, then, for sufficiently small $\eta$, $\boldsymbol{E}^\top \overline{\boldsymbol{W}}^{(k+1)}$ is invertible with probability 1. Further we will show that $\widetilde{\gamma}_i^{(\eta)}(t)$ is monotonically decreasing, which implies $\widetilde{\gamma}_i^{(\eta)}(t)$ is finite for all $t > 0$. Thus $\boldsymbol{E}^\top \overline{\boldsymbol{W}}^{(k)}$ is invertible for all $k > 0$.

It is easy to verify that $\left( \overline{\boldsymbol{W}}^{(k)}, \widetilde{\gamma}_i^{(\eta)}(\eta k) \right)_{k=1}^{n}$ form a sequence satisfying strong Markov property. By Corollary 4.2 in §7.4 of [9], if

$$b_i = \lim_{\eta \to 0} \mathbb{E} \left[ \frac{\Delta \left( \widetilde{\gamma}_i^{(\eta)2} \right)}{\eta} \,\middle|\, \boldsymbol{W} \right] \le \infty \quad \text{and} \quad \sigma_i^2 = \lim_{\eta \to 0} \mathbb{E} \left[ \frac{\left[ \Delta \left( \widetilde{\gamma}_i^{(\eta)2} \right) \right]^2}{\eta} \,\middle|\, \boldsymbol{W} \right] = 0,$$

where

$$\begin{aligned}
\Delta \left( \widetilde{\gamma}_i^{(\eta)2} \right) &= \widetilde{\gamma}_i^{(\eta)2}(t + \eta) - \widetilde{\gamma}_i^{(\eta)2}(t) \\
&= \left\| e_i^\top (\boldsymbol{W} + \Delta \boldsymbol{W}) (\boldsymbol{E}^\top \boldsymbol{W} + \boldsymbol{E}^\top \Delta \boldsymbol{W})^{-1} \right\|_2^2 - \left\| \boldsymbol{e}_i^\top (\boldsymbol{W}) (\boldsymbol{E}^\top \boldsymbol{W})^{-1} \right\|_2^2.
\end{aligned}$$

then the sequence of updates converges to the solution of the following ODE in probability

$$d\gamma_i^2(t) = b_i dt.$$

By a simple calculation,

$$\begin{aligned}
\Delta \left( \widetilde{\gamma}_i^{(\eta)2} \right) &= 2\boldsymbol{e}_i^\top \Delta \overline{\boldsymbol{W}} \left( \boldsymbol{E}^\top \overline{\boldsymbol{W}} \right)^{-1} \left( \overline{\boldsymbol{W}}^\top \boldsymbol{E} \right)^{-1} \overline{\boldsymbol{W}}^\top \boldsymbol{e}_i \\
&\quad - 2\boldsymbol{e}_i^\top \overline{\boldsymbol{W}} \left( \boldsymbol{E}^\top \overline{\boldsymbol{W}} \right)^{-1} \left( \boldsymbol{E}^\top \Delta \overline{\boldsymbol{W}} \right) \left( \boldsymbol{E}^\top \overline{\boldsymbol{W}} \right)^{-1} \left( \overline{\boldsymbol{W}}^\top \boldsymbol{E} \right)^{-1} \overline{\boldsymbol{W}} \boldsymbol{e}_i + \mathcal{O}\left( \eta^2 \left\| \overline{\boldsymbol{A}} \right\|_2^2 \right); \\
\left[ \Delta \widetilde{\gamma}_i^{(\eta)2} \right]^2 &= \mathcal{O}\left( \eta^2 \left\| \overline{\boldsymbol{A}} \right\|_2^2 \right).
\end{aligned}$$

Plugging (19) in to $\Delta\overline{\boldsymbol{W}}$, we have,

$$\begin{aligned}
&\frac{1}{\eta}\mathbb{E}\left[\Delta\left(\widetilde{\gamma}_i^{(\eta)2}\right)\middle|\overline{\boldsymbol{W}}\right] = 2\sigma_i \boldsymbol{e}_i^\top \overline{\boldsymbol{W}}\left(\boldsymbol{E}^\top\overline{\boldsymbol{W}}\right)^{-1}\left(\overline{\boldsymbol{W}}^\top\boldsymbol{E}\right)^{-1}\overline{\boldsymbol{W}}^\top\boldsymbol{e}_i \\
&\qquad - 2\boldsymbol{e}_i^\top\overline{\boldsymbol{W}}\left(\boldsymbol{E}^\top\overline{\boldsymbol{W}}\right)^{-1}\left(\boldsymbol{E}\Lambda\overline{\boldsymbol{W}}\right)\left(\boldsymbol{E}\overline{\boldsymbol{W}}\right)^{-1}\left(\overline{\boldsymbol{W}}^\top\boldsymbol{E}\right)^{-1}\overline{\boldsymbol{W}}\boldsymbol{e}_i + \mathcal{O}(\eta); \\
&= 2\sigma_i \boldsymbol{e}_i^\top \overline{\boldsymbol{W}}\left(\boldsymbol{E}^\top\overline{\boldsymbol{W}}\right)^{-1}\left(\overline{\boldsymbol{W}}^\top\boldsymbol{E}\right)^{-1}\overline{\boldsymbol{W}}^\top\boldsymbol{e}_i \\
&\qquad - 2\boldsymbol{e}_i^\top\overline{\boldsymbol{W}}\left(\boldsymbol{E}^\top\overline{\boldsymbol{W}}\right)^{-1}\left(\boldsymbol{\Lambda}_r\boldsymbol{E}^\top\overline{\boldsymbol{W}}\right)\left(\boldsymbol{E}^\top\overline{\boldsymbol{W}}\right)^{-1}\left(\overline{\boldsymbol{W}}^\top\boldsymbol{E}\right)^{-1}\overline{\boldsymbol{W}}\boldsymbol{e}_i + \mathcal{O}(\eta); \\
&\leq 2\left(\sigma_i - \sigma_r\right)\boldsymbol{e}_i^\top\overline{\boldsymbol{W}}\left(\boldsymbol{E}^\top\overline{\boldsymbol{W}}\right)^{-1}\left(\overline{\boldsymbol{W}}^\top\boldsymbol{E}\right)^{-1}\overline{\boldsymbol{W}}^\top\boldsymbol{e}_i + \mathcal{O}(\eta) \\
&= 2(\sigma_i - \sigma_r)\widetilde{\gamma}_i^2 + \mathcal{O}(\eta);
\end{aligned} \tag{20}$$

and

$$\mathbb{E}\left(\left[\Delta\left(\widetilde{\gamma}_i^{(\eta)2}\right)\right]^2\middle|\boldsymbol{W}\right) = \mathcal{O}(\eta^2). \tag{21}$$

Thus we have

$$b_i \leq 2(\sigma_i - \sigma_r)\widetilde{\gamma}_i{}^2 \quad \text{and} \quad \sigma_i = 0,$$

as desired. □

**Lemma 3.** *Suppose $\boldsymbol{W} \in \mathbb{R}^{2m\times r}$ is a matrix with $\|\boldsymbol{W}^\top\boldsymbol{W} - \boldsymbol{I}_r\|_\infty = \mathcal{O}(\delta)$. If $\left\|\overline{\boldsymbol{E}}^\top\boldsymbol{W}\right\|_{\mathrm{F}}^2 \leq \mathcal{O}(\delta)$, then*

$$\sigma_{\min}\left(\boldsymbol{W}^{(0)\top}\Lambda\boldsymbol{W}^{(0)}\right) \geq \sigma_r - \mathcal{O}(\delta).$$

## A.2 Proofs of Theorem 2

*Proof of Lemma 3.* Since $\|\boldsymbol{E}^\top\boldsymbol{W}\|_{\mathrm{F}}^2 \geq r - \mathcal{O}(\delta)$, for any $1 \leq i \leq r$, we have

$$\left\|\boldsymbol{W}\boldsymbol{e}_i'\right\|_2^2 \geq 1 - \mathcal{O}(\delta),$$

where $\boldsymbol{e}_i' \in \mathbb{R}^r$ is the $i$-th $r$-dimensional standard unit vector. Therefore,

$$\begin{aligned}
\boldsymbol{e}_i'^\top\boldsymbol{W}^\top\boldsymbol{\Lambda}\boldsymbol{W}\boldsymbol{e}_i' &= \boldsymbol{e}_i'^\top\boldsymbol{W}^\top\left(\boldsymbol{E}\boldsymbol{\Lambda}_r\boldsymbol{E}^\top + \overline{\boldsymbol{E}}\overline{\boldsymbol{\Lambda}}_r\overline{\boldsymbol{E}}^\top\right)\boldsymbol{W}\boldsymbol{e}_i' \\
&\geq \sigma_r\boldsymbol{e}_i'^\top\boldsymbol{W}^\top\boldsymbol{E}\boldsymbol{E}^\top\boldsymbol{W}\boldsymbol{e}_i' - \mathcal{O}(\delta) \\
&\geq \sigma_r - \mathcal{O}(\delta),
\end{aligned}$$

Here $\boldsymbol{\Lambda}_r = \mathrm{diag}(\sigma_1, \sigma_2, \ldots, \sigma_r)$ and $\overline{\boldsymbol{\Lambda}}_r = \mathrm{diag}(\sigma_{r+1}, \sigma_{r+2}, \ldots, \sigma_{2m})$. Hence,

$$\sigma_{\min}\left(\boldsymbol{W}^\top\boldsymbol{\Lambda}\boldsymbol{W}\right) \geq \sigma_r - \mathcal{O}(\delta).$$ □

*Proof of Theorem 2 (On independent sample).* Let

$$\zeta_{i,j} = \lim_{\eta \to 0} \eta^{-1/2} \left( \boldsymbol{e}_j'^\top \boldsymbol{\Gamma} \overline{\boldsymbol{W}}^{(\lfloor t/\eta \rfloor)\top} \boldsymbol{e}_i \right).$$

Then

$$d\zeta_{i,j} = \lim_{\eta \to 0} \Delta \zeta_{i,j}^{(\eta)}.$$

Since

$$\begin{aligned} \Delta \zeta_{i,j}^{(\eta)} &= \frac{\left( \boldsymbol{e}_j'^\top \boldsymbol{\Gamma} \Delta \boldsymbol{W} \boldsymbol{e}_i \right)}{\eta^{1/2}} \\ &= \eta^{1/2} \left[ \boldsymbol{e}_j'^\top \boldsymbol{\Gamma} \left( \overline{\boldsymbol{W}}^\top \overline{\boldsymbol{A}} - \overline{\boldsymbol{W}}^\top \overline{\boldsymbol{A} \boldsymbol{W}} \cdot \overline{\boldsymbol{W}}^\top \right) \boldsymbol{e}_i \right]. \end{aligned}$$

Therefore, we have

$$\begin{aligned} \frac{1}{\eta^{1/2}} \mathbb{E}\left( \Delta \zeta_{i,j}^{(\eta)} \middle| \boldsymbol{W} \right) &= \eta^{1/2} \mathbb{E}\left[ \boldsymbol{e}_j'^\top \left( \boldsymbol{\Gamma} \overline{\boldsymbol{W}}^\top \overline{\boldsymbol{A}} \boldsymbol{e}_i - \boldsymbol{\Gamma} \overline{\boldsymbol{W}}^\top \overline{\boldsymbol{A} \boldsymbol{W} \boldsymbol{W}}^\top \boldsymbol{e}_i \right) \right] \\ &= \eta^{1/2} \left[ \sigma_i \boldsymbol{e}_j'^\top \boldsymbol{\Gamma} \overline{\boldsymbol{W}}^\top \boldsymbol{e}_i - \boldsymbol{e}_j'^\top \boldsymbol{\Gamma} \overline{\boldsymbol{W}}^\top \boldsymbol{\Lambda} \overline{\boldsymbol{W} \boldsymbol{W}}^\top \boldsymbol{e}_i \right] \\ &= \eta^{1/2} \left[ \sigma_i \boldsymbol{e}_j'^\top \boldsymbol{\Gamma} \overline{\boldsymbol{W}}^\top \boldsymbol{e}_i - \sigma_j' \boldsymbol{e}_j'^\top \boldsymbol{\Gamma} \overline{\boldsymbol{W}}^\top \boldsymbol{e}_i \right] \\ &\leq \eta \left( \sigma_i - \sigma_r \right) \zeta_{i,j}^{(\eta)} + \mathcal{O}(\eta \delta), \end{aligned}$$

where we relies on $\boldsymbol{W}^\top \boldsymbol{W} = \boldsymbol{I}_r$ (which is true under the limit of $\eta \to 0$). We now turn to compute the infinitesimal variance $G_{i,j}$.

$$\left( \Delta \zeta_{i,j}^{(\eta)} \right)^2 = \frac{1}{\eta} \left( \boldsymbol{e}_j'^\top \boldsymbol{\Gamma} \Delta \boldsymbol{W}^\top \boldsymbol{e}_i \right)^2.$$

And

$$\mathbb{E}\left[ \left( \boldsymbol{e}_j'^\top \boldsymbol{\Gamma} \Delta \boldsymbol{W}^\top \boldsymbol{e}_i \right)^2 \middle| \boldsymbol{W} \right] = \eta^2 \mathbb{E}\left[ \left( \boldsymbol{e}_j'^\top \boldsymbol{\Gamma} \overline{\boldsymbol{W}}^\top \overline{\boldsymbol{A}} (\boldsymbol{I} - \overline{\boldsymbol{W} \boldsymbol{W}}^\top) \boldsymbol{e}_i \right)^2 \middle| \boldsymbol{W} \right].$$

We further observe that $\overline{\boldsymbol{W}} \boldsymbol{\Gamma}^\top e_j'$ is perpendicular to $(I - \overline{\boldsymbol{W} \boldsymbol{W}}^\top) e_i$ for any $r < i \leq 2m$, and that

$$\left\| (\boldsymbol{I} - \overline{\boldsymbol{W} \boldsymbol{W}}^\top) \boldsymbol{e}_i \right\| \leq 1.$$

Thus, for any $j = 1, 2, \ldots, r$,

$$\sum_{i=r+1}^{2m} \mathbb{E}\left[ \left( \boldsymbol{e}_j'^\top \boldsymbol{\Gamma} \overline{\boldsymbol{W}}^\top \overline{\boldsymbol{A}} (\boldsymbol{I} - \overline{\boldsymbol{W} \boldsymbol{W}}^\top) \boldsymbol{e}_i \right)^2 \middle| \boldsymbol{W} \right] \leq B.$$

By Corollary 4.2 in §7.4 of [9], we obtain the desired SDE with

$$K_{i,j} \leq (\sigma_i - \sigma_r) \quad \text{and} \quad \sum_{i>r}^{2m} G_{i,j}^2 \leq B,$$

which completes the proof. □

## A.3 Dependent Data

We now proceed to present the results for dependent data. Before the proof of Theorem 1 we first present the mixing lemma of Markov chain.

**Lemma 4** (Markov Chain Mixing Time, [18]). *Suppose $P \in \mathbb{R}^{m\times m}$ is the transition matrix of a Markov chain. Let $\mu$ be the stationary distribution of the chain. Denote*

$$\Phi = \min_{\Omega\subset[m]} \frac{\sum_{j\in\Omega,\ell\in\Omega^c}\sum_{i\in[m]}\frac{\mu_j p_{j,i}\mu_\ell p_{\ell,i}}{\mu_i}}{\sum_{j\in\Omega}\mu_j} \quad \textit{subject to} \quad \mu(\Omega)\le 1/2,$$

*as the merging conductance of the Markov chain. Then the mixing time, defined as*

$$\tau(\epsilon) = \min_t \left\{ t : \forall t' > t, \max_{i,j}\left|(P^{t'})_{i,j} - \mu_j\right| \le \epsilon \right\},$$

*satisfies,*

$$\tau(\epsilon) \le \frac{2}{\Phi^2}\log\left(\sqrt{\frac{\mu_{\max}}{\mu_{\min}}}\frac{1}{\epsilon}\right).$$

We now proceed with the proof of Theorem 1.

*Full Proof of Theorem 1.* Let $s^{(0)}, s^{(1)}, s^{(2)}, \ldots, s^{(k)}, \ldots s^{(n)}$ be samples from the Markov chain. We next show that the statements in Lemma 1 and Lemma 2 are true for *Markov Samples*: $s^{(\tau-1)}, s^{(\tau)}, s^{(2\tau-1)}, s^{(2\tau)}, \ldots, s^{(k\tau-1)}, s^{(k\tau)}\ldots$, where $\tau = \tau(\eta)$. We start with Lemma 1. We first observe that the updates of the Algorithm combining with Markov samples satisfy strong Markov property. Denote

$$\boldsymbol{A}^{(i)} = \begin{bmatrix} 0 & \boldsymbol{Z}^{(k)} \\ \boldsymbol{Z}^{(k)\top} & 0 \end{bmatrix} \in \mathbb{R}^{2m\times 2m}.$$

Assumption 1 and Assumption 2 can be verified, conditioning on $s^{(0)}$. By definition of $\tau = \tau(\eta)$, we also observe that,

$$\mathbb{E}\left(A^{(k)}\middle|s^{(0)}\right) = \begin{bmatrix} 0 & \boldsymbol{D}\boldsymbol{P} + \widehat{\boldsymbol{E}}\boldsymbol{P} \\ \boldsymbol{P}^\top\boldsymbol{D} + \boldsymbol{P}^\top\widehat{\boldsymbol{E}} & 0 \end{bmatrix} \in \mathbb{R}^{2m\times 2m}.$$

where $\widehat{\boldsymbol{E}}$ is a diagonal matrix with

$$\forall i \in [m] : \left|\widehat{\boldsymbol{E}}_{i,i}\right| \le \eta.$$

Therefore, the spectrum of $\mathbb{E}\left(\boldsymbol{A}^{(k)}|s^{(0)}\right)$ differs with independent case by at most an additive $\mathcal{O}(\eta)$ term. Thus Equation (20) holds, hence the rest of the proof of Theorem 1 is exactly the same as proof of the independent case.

Theorem 2 follows similarly by observing that the spectrum of $\mathbb{E}\left(\boldsymbol{A}^{(i)}\right)$ differs with independent case by at most an additive $\mathcal{O}(\eta)$ term.

The rest of proof of this theorem is the same as that of Theorem 1. □

### A.4 Proof of Theorem 3

*Proof.* Suppose we have

$$\left\|\overline{\boldsymbol{V}}^\top \widehat{\boldsymbol{V}}\right\|_{\mathrm{F}}^2 = \sum_{j,j'=1}^{r}\left(\sum_{i=1}^{m} \boldsymbol{v}_{i,j}\widehat{\boldsymbol{v}}_{i,j'}\right)^2 = \sum_{i,i'=1}^{m}\left(\sum_{j=1}^{r} \boldsymbol{v}_{i,j}\widehat{\boldsymbol{v}}_{i',j}\right)^2 \geq r-\epsilon \quad \text{and}$$
$$\widehat{\mu_i} \in (1 \pm \epsilon')\mu_i.$$

Let

$$\boldsymbol{V}^\top \widehat{\boldsymbol{V}} = \widetilde{\boldsymbol{U}}^\top \boldsymbol{T} \widetilde{\boldsymbol{U}}',$$

where $\boldsymbol{T} = \mathrm{diag}(\widehat{\sigma_1}, \widehat{\sigma_2}, \dots, \widehat{\sigma_r})$ and $\widetilde{\boldsymbol{U}}, \widetilde{\boldsymbol{U}}'$ are orthonormal matrices. Since for each $i$, $|\widehat{\sigma_i}| \leq 1$, we have

$$r - \epsilon \leq \sum_{i=1}^{r} \widehat{\sigma}_i^2 \leq r.$$

Fixing $\sum_{i=1}^r \widehat{\sigma}_i^2$, the minimum value of $\sum_{i=1}^r \widehat{\sigma_i}$ is obtained when $\widehat{\sigma_1} = \widehat{\sigma_2} = \dots \widehat{\sigma}_{r-1} = 1$ and $\widehat{\sigma_r} = 1 - \epsilon$. Let $\widetilde{\boldsymbol{V}} = \boldsymbol{V}\widetilde{\boldsymbol{U}}^\top$ and $\widetilde{\widehat{\boldsymbol{V}}} = \widehat{\boldsymbol{V}}\widetilde{\boldsymbol{U}}'^\top$. We have

$$\sum_{i=1}^{m} \widetilde{\boldsymbol{V}}_{i*}^\top \widetilde{\widehat{\boldsymbol{V}}}_{i*} = \sum_{i=1}^{m}\sum_{j=1}^{r} \widetilde{\boldsymbol{V}}_{i,j} \widetilde{\widehat{\boldsymbol{V}}}_{i,j} = \sum_{j=1}^{r} \widetilde{\boldsymbol{V}}_{*j}^\top \widetilde{\widehat{\boldsymbol{V}}}_{*j} = \sum_{j=1}^{r} \widehat{\sigma_j} \in [r-\epsilon, r]. \tag{22}$$

For any $s_i, s_j \in S$, the distance

$$d(s_i, s_j) = \left\|\frac{\boldsymbol{V}_{i*}}{\mu_i} - \frac{\boldsymbol{V}_{j*}}{\mu_j}\right\| = \left\|\frac{\widetilde{\boldsymbol{V}}_{i*}}{\mu_i} - \frac{\widetilde{\boldsymbol{V}}_{j*}}{\mu_j}\right\|$$
$$d'(s_i, s_j) = \left\|\frac{\widehat{\boldsymbol{V}}_{i*}}{\widehat{\mu_i}} - \frac{\widehat{\boldsymbol{V}}_{j*}}{\widehat{\mu_j}}\right\| = \left\|\frac{\widetilde{\widehat{\boldsymbol{V}}}_{i*}}{\widehat{\mu_i}} - \frac{\widetilde{\widehat{\boldsymbol{V}}}_{j*}}{\widehat{\mu_j}}\right\|.$$

Therefore, by triangle inequality,

$$d'(s_i, s_j) = \left\|\frac{\widetilde{\widehat{\boldsymbol{V}}}_{i*}}{\widehat{\mu_i}} - \frac{\widetilde{\boldsymbol{V}}_{i*}}{\mu_i} + \frac{\widetilde{\boldsymbol{V}}_{i*}}{\mu_i} - \frac{\widetilde{\widehat{\boldsymbol{V}}}_{j*}}{\widehat{\mu_j}} + \frac{\widetilde{\boldsymbol{V}}_{j*}}{\mu_j} - \frac{\widetilde{\boldsymbol{V}}_{j*}}{\mu_j}\right\|$$
$$\in \left\|\frac{\widetilde{\boldsymbol{V}}_{i*}}{\mu_i} - \frac{\widetilde{\boldsymbol{V}}_{j*}}{\mu_j}\right\| \pm \left\|\frac{\widetilde{\widehat{\boldsymbol{V}}}_{i*}}{\widehat{\mu_i}} - \frac{\widetilde{\boldsymbol{V}}_{i*}}{\mu_i} - \frac{\widetilde{\widehat{\boldsymbol{V}}}_{j*}}{\widehat{\mu_j}} + \frac{\widetilde{\boldsymbol{V}}_{j*}}{\mu_j}\right\|.$$

Now consider the second term,

$$\left\|\frac{\widetilde{\widehat{\boldsymbol{V}}}_{i*}}{\widehat{\mu_i}} - \frac{\widetilde{\boldsymbol{V}}_{i*}}{\mu_i} - \frac{\widetilde{\widehat{\boldsymbol{V}}}_{j*}}{\widehat{\mu_j}} + \frac{\widetilde{\boldsymbol{V}}_{j*}}{\mu_j}\right\| \leq \left\|\frac{\widetilde{\widehat{\boldsymbol{V}}}_{i*}}{\widehat{\mu_i}} - \frac{\widetilde{\boldsymbol{V}}_{i*}}{\mu_i}\right\| + \left\|-\frac{\widetilde{\widehat{\boldsymbol{V}}}_{j*}}{\widehat{\mu_j}} + \frac{\widetilde{\boldsymbol{V}}_{j*}}{\mu_j}\right\|$$

Since we have $\widehat{\mu_i} = (1 \pm \epsilon')\mu_i$, where $\epsilon' = \sqrt{\epsilon}$,

$$\left\| \frac{\widehat{\widetilde{\boldsymbol{V}}}_{i*}}{\widehat{\mu_i}} - \frac{\widetilde{\boldsymbol{V}}_{i*}}{\mu_i} - \frac{\widehat{\widetilde{\boldsymbol{V}}}_{j*}}{\widehat{\mu_j}} + \frac{\widetilde{\boldsymbol{V}}_{j*}}{\mu_j} \right\|^2 \leq \frac{16}{\mu_i^2} \left\| \widehat{\widetilde{\boldsymbol{V}}}_{i*} - \widetilde{\boldsymbol{V}}_{i*} \right\|^2 + \frac{16}{\mu_j^2} \left\| \widehat{\widetilde{\boldsymbol{V}}}_{j*} - \widetilde{\boldsymbol{V}}_{j*} \right\|^2 + \frac{16\epsilon'^2}{\mu_i^2} \left\| \widehat{\widetilde{\boldsymbol{V}}}_{i*} \right\|^2 + \frac{16\epsilon'^2}{\mu_j^2} \left\| \widehat{\widetilde{\boldsymbol{V}}}_{j*} \right\|^2 .$$

Observe that, for any $i \in [n]$,

$$\left\| \widehat{\widetilde{\boldsymbol{V}}}_{i*} - \widetilde{\boldsymbol{V}}_{i*} \right\|^2 = \left\| \widehat{\widetilde{\boldsymbol{V}}}_{i*} \right\|^2 + \left\| \widetilde{\boldsymbol{V}}_{i*} \right\|^2 - 2\widehat{\widetilde{\boldsymbol{V}}}_{i*}^{\top} \widetilde{\boldsymbol{V}}_{i*}.$$

Therefore

$$\sum_{i=1}^{m} \left\| \widehat{\widetilde{\boldsymbol{V}}}_{i*} - \widetilde{\boldsymbol{V}}_{i*} \right\|^2 = 2r - 2\sum_{i=1}^{m} \widehat{\widetilde{\boldsymbol{V}}}_{i*}^{\top} \widetilde{\boldsymbol{V}}_{i*} = 2r - 2\sum_{j=1}^{r} \widehat{\sigma_j} \leq 2\epsilon.$$

Thus, for each $i \in [n]$

$$\left\| \widehat{\widetilde{\boldsymbol{V}}}_{i*} - \widetilde{\boldsymbol{V}}_{i*} \right\|^2 \leq 2\epsilon.$$

Furthermore, since

$$\epsilon'^2 \leq \epsilon$$

we have

$$\left| d'^2(s_i, s_j) - d^2(s_i, s_j) \right| \leq \frac{96\epsilon}{\mu_{\min}^2}.$$

□

# B Use Geometric Median to Boost The Probability of Success

In this section we show that the geometric median indeed boosts the probability of success. In particular, let $\overline{\boldsymbol{V}} \in \mathbb{R}^{2m \times r}$ be a fixed matrix with orthonormal columns. Let $k = \mathcal{O}\left(\log \frac{1}{\delta}\right)$ and $\boldsymbol{V}_1, \boldsymbol{V}_2, \ldots \boldsymbol{V}_k$ be independent random column-orthonormal matrices from $\mathbb{R}^{2m \times r}$ with

$$\forall i \in [k] : \mathbb{P}\left[ \left\| \boldsymbol{V}_i^{\top} \overline{\boldsymbol{V}} \right\|_{\mathrm{F}}^2 < r - \epsilon \right] \leq \frac{1}{4}.$$

We denote $\boldsymbol{H}$ be the geometric median of $\boldsymbol{V}_1 \boldsymbol{V}_1^{\top}, \boldsymbol{V}_2 \boldsymbol{V}_2^{\top}, \ldots, \boldsymbol{V}_k \boldsymbol{V}_k^{\top}$ in Euclidean space, i.e.

$$\boldsymbol{H} := \underset{\boldsymbol{Z} \in \mathbb{R}^{2m \times 2m}}{\operatorname{argmin}} \sum_{j=1}^{k} \left\| \boldsymbol{V}_j \boldsymbol{V}_j^{\top} - \boldsymbol{Z} \right\|_{\mathrm{F}}.$$

Let

$$\widetilde{\boldsymbol{V}} = \underset{\boldsymbol{V}_i : i \in [k]}{\operatorname{argmin}} \left\| \boldsymbol{V}_i \boldsymbol{V}_i^\top - \boldsymbol{H} \right\|_{\mathrm{F}}.$$

The following proposition guarantees $\widetilde{\boldsymbol{V}}$ to be a good estimation of $\overline{\boldsymbol{V}}$.

**Proposition 4.**

$$\mathbb{P}\left[ \left\| \widetilde{\boldsymbol{V}}^\top \overline{\boldsymbol{V}} \right\|_{\mathrm{F}}^2 < r - C\epsilon \right] \leq \delta,$$

*for some absolute constant $C$.*

*Proof.* We first show that the quantity $\left\| \boldsymbol{V}_i^\top \overline{\boldsymbol{V}} \right\|_{\mathrm{F}}^2$ is related to Euclidean distance in $\mathbb{R}^{2m \times 2m}$:

$$\left\| \boldsymbol{V}_i^\top \overline{\boldsymbol{V}} \right\|_{\mathrm{F}}^2 = \operatorname{tr}\left( \boldsymbol{V}_i^\top \overline{\boldsymbol{V}\boldsymbol{V}}^\top \boldsymbol{V}_i \right) = \operatorname{tr}\left( \boldsymbol{V}_i \boldsymbol{V}_i^\top \overline{\boldsymbol{V}\boldsymbol{V}}^\top \right) = r - \frac{\left\| \boldsymbol{V}_i \boldsymbol{V}_i^\top - \overline{\boldsymbol{V}\boldsymbol{V}}^\top \right\|_{\mathrm{F}}^2}{2}.$$

Thus we obtain,

$$\forall i \in [k] : \mathbb{P}\left[ \left\| \boldsymbol{V}_i \boldsymbol{V}_i^\top - \overline{\boldsymbol{V}\boldsymbol{V}}^\top \right\|_{\mathrm{F}}^2 \geq 2\epsilon \right] \leq \frac{1}{4}.$$

Then by Theorem 3.1 of [19],

$$\mathbb{P}\left[ \left\| \boldsymbol{H} - \overline{\boldsymbol{V}\boldsymbol{V}}^\top \right\|_{\mathrm{F}}^2 \geq C_1 \epsilon \right] \leq \frac{\delta}{2}$$

for some absolute constant $C_1$ and appropriately chosen $k$. Since the $\boldsymbol{V}_i$s are independent, we obtain

$$\mathbb{P}\left[ \forall i \in [k] : \left\| \boldsymbol{V}_i \boldsymbol{V}_i^\top - \overline{\boldsymbol{V}\boldsymbol{V}}^\top \right\|_{\mathrm{F}}^2 \geq 2\epsilon \right] \leq \frac{1}{4^k} \leq \frac{\delta}{2},$$

for appropriately chosen $k$. Let $\mathcal{E}$ be the following event,

$$\left\| \boldsymbol{H} - \overline{\boldsymbol{V}\boldsymbol{V}}^\top \right\|_{\mathrm{F}}^2 < C_1 \epsilon \quad \text{and} \quad \exists i^* : \left\| V_{i^*} V_{i^*}^\top - \overline{\boldsymbol{V}\boldsymbol{V}}^\top \right\|_{\mathrm{F}}^2 < 2\epsilon.$$

By union bound, $\mathbb{P}[\mathcal{E}] \geq 1 - \delta$. For the rest of the proof, we condition on $\mathcal{E}$. Thus

$$\begin{aligned}
\left\| \widetilde{\boldsymbol{V}}\widetilde{\boldsymbol{V}}^\top - \overline{\boldsymbol{V}\boldsymbol{V}}^\top \right\|_{\mathrm{F}} &= \left\| \widetilde{\boldsymbol{V}}\widetilde{\boldsymbol{V}}^\top - \boldsymbol{H} + \boldsymbol{H} - \overline{\boldsymbol{V}\boldsymbol{V}}^\top \right\|_{\mathrm{F}} \\
&\leq \left\| \widetilde{\boldsymbol{V}}\widetilde{\boldsymbol{V}}^\top - \boldsymbol{H} \right\|_{\mathrm{F}} + \left\| \overline{\boldsymbol{V}\boldsymbol{V}}^\top - \boldsymbol{H} \right\|_{\mathrm{F}} \\
&\leq \left\| \boldsymbol{V}_{i^*} \boldsymbol{V}_{i^*}^\top - \boldsymbol{H} \right\|_{\mathrm{F}} + \sqrt{C_1 \epsilon} \\
&\leq \sqrt{C_1 \epsilon} + \left\| \boldsymbol{V}_{i^*} \boldsymbol{V}_{i^*}^\top - \overline{\boldsymbol{V}\boldsymbol{V}}^\top \right\|_{\mathrm{F}} + \left\| \overline{\boldsymbol{V}\boldsymbol{V}}^\top - \boldsymbol{H} \right\|_{\mathrm{F}} \\
&\leq \left( 2\sqrt{C_1} + \sqrt{2} \right) \sqrt{\epsilon}.
\end{aligned}$$

Thus

$$\left\| \widetilde{\boldsymbol{V}}\widetilde{\boldsymbol{V}}^\top - \overline{\boldsymbol{V}\boldsymbol{V}}^\top \right\|_{\mathrm{F}}^2 \leq \left( 2\sqrt{C_1} + \sqrt{2} \right)^2 \epsilon \Rightarrow \left\| \boldsymbol{V}_i^\top \overline{\boldsymbol{V}} \right\|_{\mathrm{F}}^2 \geq r - \frac{\left( 2\sqrt{C_1} + \sqrt{2} \right)^2}{2} \epsilon.$$

The proof is completed by setting $C = \frac{\left(2\sqrt{C_1} + \sqrt{2}\right)^2}{2}$. □

Moreover, $\boldsymbol{H}$ can be obtained efficiently by running the algorithm presented in [8].